\documentclass[runningheads]{llncs}
\usepackage{graphicx}
\usepackage{subcaption}
\usepackage{multirow}
\usepackage{booktabs}
\usepackage{tabularx}

\begin{document}
\title{Fostering Event Compression using \\ Gated Surprise}
\titlerunning{Fostering Event Compression using Gated Surprise}
%
%
\author{Dania Humaidan\thanks{Supported by International Max Planck Research School for Intelligent Systems.}\orcidID{0000-0003-1381-257X} \and
Sebastian Otte\orcidID{0000-0002-0305-0463} \and
Martin V. Butz\orcidID{0000-0002-8120-8537}}
\authorrunning{D. Humaidan et al.}
%
\institute{University of Tuebingen - Neuro-Cognitive Modeling Group, Sand 14, 72076 Tuebingen - Germany}
\maketitle              
\begin{abstract}
Our brain receives a dynamically changing stream of sensorimotor data. 
Yet, we perceive a rather organized world, which we segment into and perceive as events. 
Computational theories of cognitive science on event-predictive cognition suggest that our brain forms generative, event-predictive models by segmenting sensorimotor data into suitable chunks of contextual experiences. 
Here, we introduce a hierarchical, surprise-gated recurrent neural network architecture, which models this process and develops compact compressions of distinct event-like contexts. 
The architecture contains a contextual LSTM layer, which develops generative compressions of ongoing and subsequent contexts. 
These compressions are passed into a GRU-like layer, which uses surprise signals to update its recurrent latent state. 
The latent state is passed forward into another LSTM layer, which processes actual dynamic sensory flow in the light of the provided latent, contextual compression signals. 
Our model shows to develop distinct event compressions and achieves the best performance on multiple event processing tasks. 
The architecture may be very useful for the further development of resource-efficient learning, hierarchical model-based reinforcement learning, as well as the development of artificial event-predictive cognition and intelligence.

\end{abstract}
\section{Introduction}
The way our brain perceives information and organizes it remains an open question. 
It appears that we have a tendency to perceive, interpret, and thus understand our sensorimotor data streams in the form of events. 
The so-called Event Segmentation Theory (EST) \cite{zacks_event_2007} suggests that we utilize temporary increases in prediction errors for segmenting the stream of sensorimotor information into separable events \cite{franklin_structured_2019}.
As a result, compact event encodings develop. 

Event encodings certainly do not develop for their own sake or for the mere sake of representing sensorimotor information, though. 
Rather, it appears that event encodings are very useful for memorizing events and event successions, as well as for enabling effective hierarchical reinforcement learning \cite{Botvinick:2009}, amongst other benefits \cite{Butz:2016}.
Indeed it appears that our brain prepares for upcoming events in the prefrontal cortex \cite{tanji_behavioral_2001}. 
Moreover, the whole midline default network \cite{Stawarczyk:2019tsi} seems to actively maintain a push-pull relationship between externally-generated stimulations and internally-generated imaginations, including retrospective reflections and prospective anticipations.

We have previously modeled such processes with REPRISE -- a retrospective and prospective inference scheme \cite{Butz:2019,Butz:2019ICANN} -- showing promising planning and event-inference abilities.
However, the retrospective inference mechanism is also rather computationally expensive. 
Here, we introduce a complementary surprise-processing modular architecture, which may support the event-inference abilities of REPRISE as well as, more generally speaking, the development of event-predictive compressions. 
We show that when contextual information is selectively routed into a predictive processing layer via GRU-like \cite{Chung:2014} switching-gates, suitable event compressions are learned via standard back-propagation through time. 
As a result, the architecture can generate and instantly switch between distinct functional operations. 

After providing some further background on related event processing mechanisms in humans and neural modeling approaches, we introduce our surprise-processing modular architecture.
We evaluate our system exemplarily on a set of simple function prediction tasks, where the lower-layer network needs to predict function value outputs given inputs and contextual information.
Meanwhile, deeper layers learn to distinguish different functional mappings, compressing the individual functions into event-like encodings.
In the near future, we hope to scale this system to more challenging real-world tasks and to enhance the architecture such that upcoming surprise signals and consequent event switches are anticipated as well.

\section{Related Work}

The ability to distinguish different contexts was previously tested in humans \cite{zacks_event_2007,zhao_perception_2014,serrano_movie_2017}.
Segmenting these events was suggested to make use of the prediction failures to update the internal model and suggest that a new event has begun.
Loschky and colleagues \cite{loschky_what_2015} showed a group of participants selected parts of a film. They showed that when the clip could be put within a larger context, the participant had more systematic eye movements.
Baldassano and colleagues \cite{baldassano_representation_2018} showed that the participants had consistently different brain activity patterns for different ongoing contexts (flying from the airport and eating at a restaurant).
Pettijohn and colleagues have shown that increasing the number of event boundaries can have a positive effect on memory \cite{pettijohn_event_2016}.

From the computational aspect, the usage of prediction error to predict the next stimulus was presented in the work of Reynolds and colleagues \cite{reynolds_computational_2007} who used a feed forward network in combination with a recurrent neural network module, memory cells, and a gating mechanism. This model was later extended with an RL agent that controls the gating mechanism with a learned policy \cite{metcalf_modeling_nodate}. Successfully segmenting the information stream into understandable units was also attempted with reservoir computing \cite{asabuki_interactive_2018}. It was shown that this mechanism can be sufficient to identify event boundaries.
However, it did not develop a hierarchical structure that is believed to be present when considering action production and goal directed behaviors \cite{koechlin_architecture_2003}.

A framework that includes a hierarchical system of multilevel control was illustrated in \cite{li_combined_2019}, which offers a background survey and a general hierarchical framework for neuro-robotic systems (NRS).
In this framework, the processing of the perceptual information happens in a high level cognition layer whose output passes through a translational mid level layer to a lower level execution layer. The lower layer includes the sensory feedback between the agent and the surrounding environment.
An interesting hierarchical structure for spatial cognition was presented in the work of Martinet and colleagues \cite{martinet_spatial_2011}. Their presented model showed how interrelated brain areas can interact to learn how to navigate towards a target by investigating different policies to find the optimal one. However, this structure focused on a certain maze and only used the size of the reward at the goal location to make decisions.

Another important aspect of forming loosely hierarchical structured event compressions lies in the prediction of event boundaries. 
Indeed, it was shown that having background knowledge about the ongoing activities while an event unfolds can help to predict when the current event might end \cite{hard_attention_2019}.
This means that the developing event-generative compression structure may develop deeper knowledge about the currently unfolding event.
Amongst other things, such structures may develop event boundary anticipating encodings, which, when activated, predict initially unpredictable event changes.

\section{Surprise-Processing Modular Architecture}
We present a hierarchical surprise-gated recurrent neural network architecture.
The system simulates the flow of contextual information from a deep layer, which prepares the processing of upcoming events developing event-compressing encodings.
These encodings are used to modify the processing of the lower-level sensor- or sensorimotor-processing layer.
In between, a GRU-like gating layer controls when novel context modifying signals are passed on to the lower-level layer and when the old signal should be kept.
As a result, the lower-level layer is generating predictions context-dependently, effectively learning to distinguish different events.

\subsection{The Structure}
\begin{figure}[t!]
	\centering
\includegraphics[width=.9\textwidth, height=6.15cm]{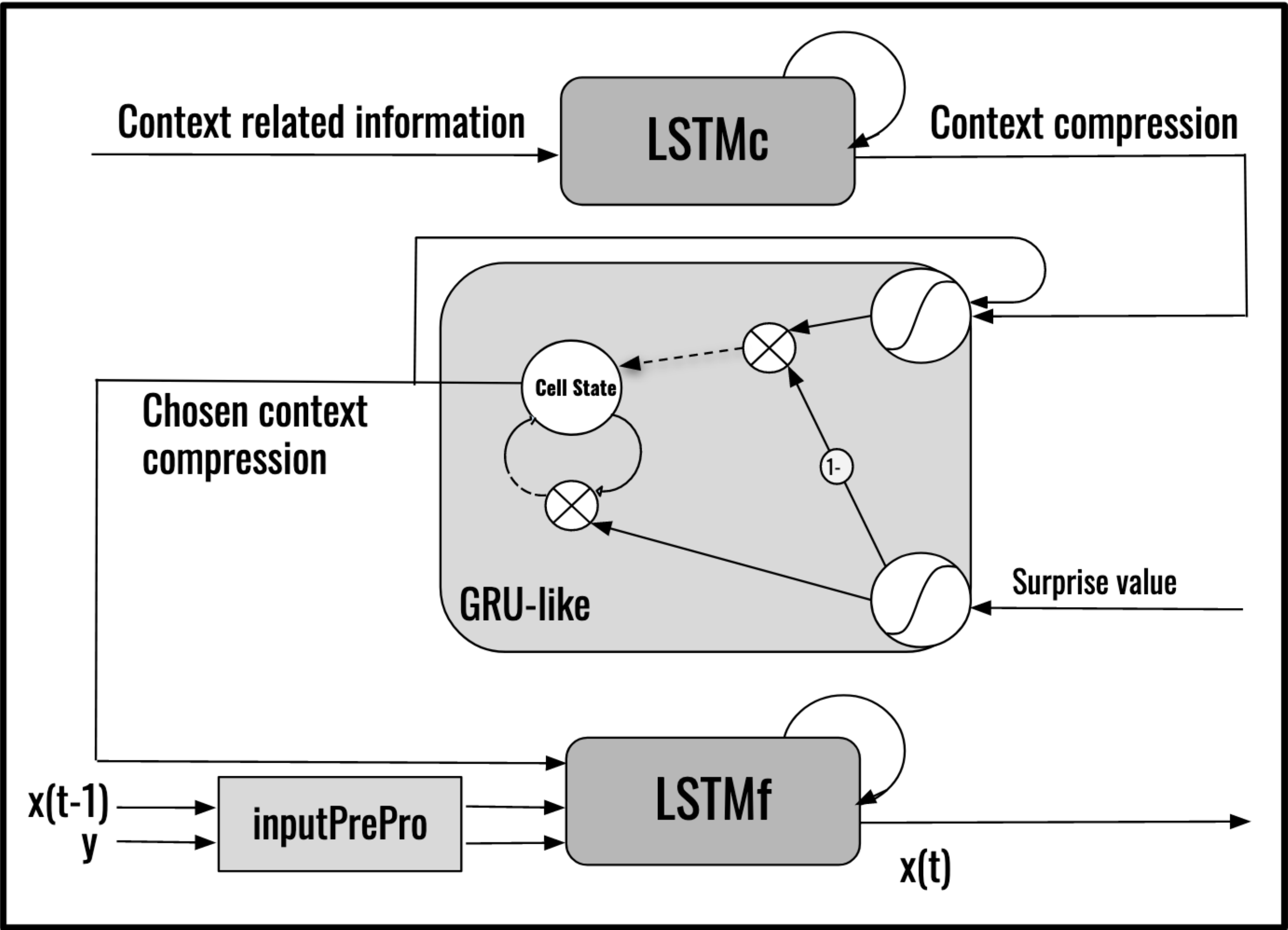}
\caption{The hierarchical structure composed of a deep contextual layer (LSTMc), a GRU-like gating layer and a low-level function processing layer (LSTMf). Additionally, we added an MLP to preprocess the function input (inputPrePro).}

 \label{fig:structure}
\end{figure}

The proposed structure is composed of a contextual recurrent neural network layer, implemented using an LSTM network (LSTMc) \cite{Hochreiter:1997}, which is responsible for generating an event compression that is representing the currently ongoing or next upcoming context.
This contextual information is fed into a middle layer, which is implemented by a GRU-like gate \cite{Chung:2014}. 
The gate decides how much of the novel contextual information in proportion to the previous contextual information will be passed on to the lower level layer.
This lower level function processing layer, which is also implemented by an LSTM, predicts a function value (LSTMf).
The function input is preprocessed using an MLP unit (inputPrePro), before being provided to LSTMf. 
The structure is shown in Figure~\ref{fig:structure}.
Note that the dotted lines denote unweighted inputs.

The decision about the current context is taken at the GRU-like top down gating layer. 
When a new event begins, LSTMf will produce erroneous predictions as the function switched.
As a result, this correspondingly large surprise value, representing the unexpectedly high prediction error \cite{Butz:2019}, may be provided to the gating layer.
A surprise signal can thus be used to manipulate the update gate of a GRU layer, receiving and passing on new contextual information from LSTMc surprise-dependently. 
If the context has not changed, then the gate stays closed, and the same old event compression is provided as the contextual information to the LSTMf layer.

\subsection{The Switch GRU}
The used GRU structure was adapted to act as a switch to decide when (i) to keep the gate closed, in which case the already saved context compression from the previous time step will be maintained and passed on, or (ii) to open the gate, in which case the new context generated by LSTMc will flow in.
To perform this task, the update gate at the GRU is modified to be unweighted, with its input being the surprise signal.
The combined gate is now getting the new context compression from LSTMc and the hidden cell state (context compression of the previous time step) inputs.
The reset gate is removed as it has no role here.

\section{Experiments and Results}
For evaluation purposes, we used an example of a time series that includes four functions $f_e(x,y)$ representing four different contexts or events $e$.
Converting this into a continuous time series, the previous function output is used as the first function input at the following time step $t$, that is, $x_t=f_e(x_{t-1},y_{t-1})$. 
Meanwhile, function inputs $y_t$ are generated independently and uniformly distributed between $-1$ and $1$. 
The four functions are
\begin{enumerate}
	\item An addition function (add): $f_{add}(x,y) = 0.9x+y$,
	\item A sine function (sin): $f_{sin}(x,y)= x + sin(\pi y)$.
	\item A subtraction function (sub): $f_{sub}(x,y) = 0.9x-y$, 
	\item A constant function (con): $f_{con}(x,y)=x$,
\end{enumerate}
Function switches occurred uniformly randomly every $5$ to $12$ times steps.

\subsection{Single Network Experiments}
As an upper error baseline, we first evaluated the performance of a single LSTM layer or a two-layer perceptron (MLP), which receives $x$ and $y$ as input and is trained to learn $f_e{x,y}$ without providing any direct information about $e$.
Next, as a lower error baseline, we evaluate the performance of the two networks when we augment the input with a one-hot vector denoting the ongoing event.
Finally, to make the problem harder again and enforce the anticipation of an event switch, we switched the event information at a time point uniformly randomly earlier than the actual next scheduled event switch, but at least two time steps after the last event switch. 
This was done to simulate the idea of preparing for the next event switch before it actually happens. 
In addition, we distinguish between runs in which the consecutive functions were in the same order and runs in which the next function type $e \in \{add,sub,con,sin\}$ is randomly chosen each time.

The used LSTM network had 10 hidden units and the MLP had two hidden layers each with 50 units.
The weights of the networks were updated at a fixed rate every 20 time steps.
We used a learning rate of 10\textsuperscript{-4} and trained every network for 2\ 000 epochs each with 2\ 000 steps.
Finally, we tested every network for 150 test iterations.
Reported performance results are averaged over ten differently initialized networks.

The results are shown in Table~\ref{Tab:SingleLSTM_MLP}.
As expected, worst performance is obtained when the network does not receive any context-related information, while best performance is achieved when context information is provided.
When the order of the successive functions is randomized, performance only slightly degrades.
When the context information switches earlier than the actual function, performance degrades, yielding an average error between the case when no context information is provided and when context information is provided perfectly in tune with the actual context switches.

\begin{table}[t
	b!]
\caption{Average training prediction error in the different single LSTM layer experiments.}
\label{Tab:SingleLSTM_MLP}
\begin{scriptsize}
{
\setlength\tabcolsep{0.4pt}
\newcolumntype{Y}{>{\centering\arraybackslash}X}
\begin{tabularx}{\linewidth}{Y c c c c}
    \toprule
    \multirow{2} * {Experiment} & \multicolumn {2}{p{3.5cm}}{\centering LSTM} & \multicolumn{2}{p{3.5cm}}{\centering MLP} \\ \cmidrule(l{0pt}r{5pt}){2-3} \cmidrule(l{-2pt}r{0pt}){4-5}
    & avg. error & stdev. & avg. error  & stdev. \\
	\midrule
    No CI provided & 0.2670   & 0.0272 & 0.4180 & 0.0016 \\
    CI provided with fixed function order &  0.0533 & 0.0292 &  0.0098  & 0.0011 \\
    CI provided with random function order & 0.0551 & 0.0215 & 0.0139  & 0.0022 \\
    CI provided but switched earlier & 0.1947  &  0.0134 & 0.3180 & 0.0012\\
    \bottomrule
  \end{tabularx}
}
\end{scriptsize}
\end{table}

When comparing the performance of the LSTM with the MLP, several observations can be made.
First, when no context information or ill-tuned context information is provided, LSTM outperforms the MLP.
This is most likely the case because the LSTM can in principle infer the function that currently applies by analyzing the successive input signals.
As a result, it appears to decrease its prediction error via its recurrent information processing ability.
On the other hand, when perfect context information is provided, the MLP learns to approximate the function even better than the LSTM module, indicating that the MLP can play out its full capacity, while the recurrent connections are somewhat prohibiting better performance with the LSTM module.

\subsection{Full Network Experiments}

Next, we performed the experiments using the introduced surprise-processing modular neural architecture. 

We evaluated the structure by testing four cases:
\begin{itemize}
    \item The gate is always closed: The GRU-like layer output is constantly zero (approximately corresponding to the upper error baseline).
    \item The gate is always open: The GRU-like layer output is continuously controlled by the new context compression from LSTMc.
    \item The gate is only open at context switches: The GRU-like layer output maintains the context prediction generated by LSTMc when the context is switched.
    \item The gate is gradually open at context switches: The GRU-like layer switches its context output more smoothly.
\end{itemize}
Note that the fourth scenario is meant to probe whether a gradual surprise signal can help to predict the switches between the contexts in a smoother manner.
The gate in this case turns from being closed, to being half-open, to being fully opened, and back to half-open and closed. 

Final test errors -- again averaged over ten independently randomly weight-initialized networks - are shown in Table~\ref{Tab:PredErr+distances}.
The results show that the best results are obtained by keeping the gate closed while the same context is progressing, and only opening it when a new event starts.
As expected, the worst performance is achieved when the gate is always closed.
Note also that the performance only slightly improves when the gate is always open, indicating that the architecture cannot detect the event switches on its own. 
Gradually opening and closing the gate slightly degrades the performance in comparison to when the gate is only open at the actual switch.
When considering the differences in the compression codes that are passed down to LSTMf in the different events, the largest distances are generated by the network when the GRU-like update gate is open at the switch, only, thus indicating that it generated the most distinct compressions for the four function events.  

\begin{table}[tb!]
\caption{Average training prediction error and average distance between the centers of the clusters formed by the context compressions' values in different gate states. The lowest average error and largest average distances are marked in bold.}\label{Tab:PredErr+distances}
\begin{scriptsize}
{
\setlength\tabcolsep{0.4pt}
\newcolumntype{Y}{>{\centering\arraybackslash}X}
\begin{tabularx}{\linewidth}{ c Y Y Y Y Y }
    \toprule
    Gate status & avg. error & stdev error & Compared clusters & avg. distance& stdev distance \\  
    \midrule
Always closed& 0.280 & 0.059 & Any & 0.0 & 0.0 \\
    \midrule
     \multirow{6}{*}{Always open} & \multirow{6}{*}{0.206}& \multirow{6}{*}{0.014} &Add Sin& 0.28&0.17\\ &&&Add Sub&1.22&0.42\\&&&Add Const&0.64&0.19\\&&&Sin Sub&1.27&0.34\\&&&Sin Const&0.70&0.24\\&&&Sub Const&0.7&0.26\\
    \midrule
\multirow{6}{*}{Only open at switch} & \multirow{6}{*}{\textbf{0.059}}& \multirow{6}{*}{0.017}&Add Sin& \textbf{0.69}&0.15\\ &&&Add Sub&\textbf{3.12}&0.42\\&&&Add Const&\textbf{1.46}&0.27\\&&&Sin Sub&\textbf{2.59}&0.47\\&&&Sin Const& 0.92 &0.17\\&&&Sub Const&\textbf{1.72}&0.4\\
    \midrule
    \multirow{6}{*}{Gradually opened} & \multirow{6}{*}{0.083}& \multirow{6}{*}{0.030}&Add Sin& {0.61}&0.17\\ &&&Add Sub& 2.17 &0.69 \\&&& Add Const & 1.35 & 0.56 \\&&& Sin Sub & 1.81 & 0.39 \\&&& Sin Const & \textbf{1.00} & 0.20 \\ &&& Sub Const & 0.82 & 0.31 \\
	\bottomrule
  \end{tabularx}
}
\vspace{-.5cm}
\end{scriptsize}
\end{table}

\begin{figure}[thb!]
	\centering
	\includegraphics[width=.7\textwidth, height=4.5cm]{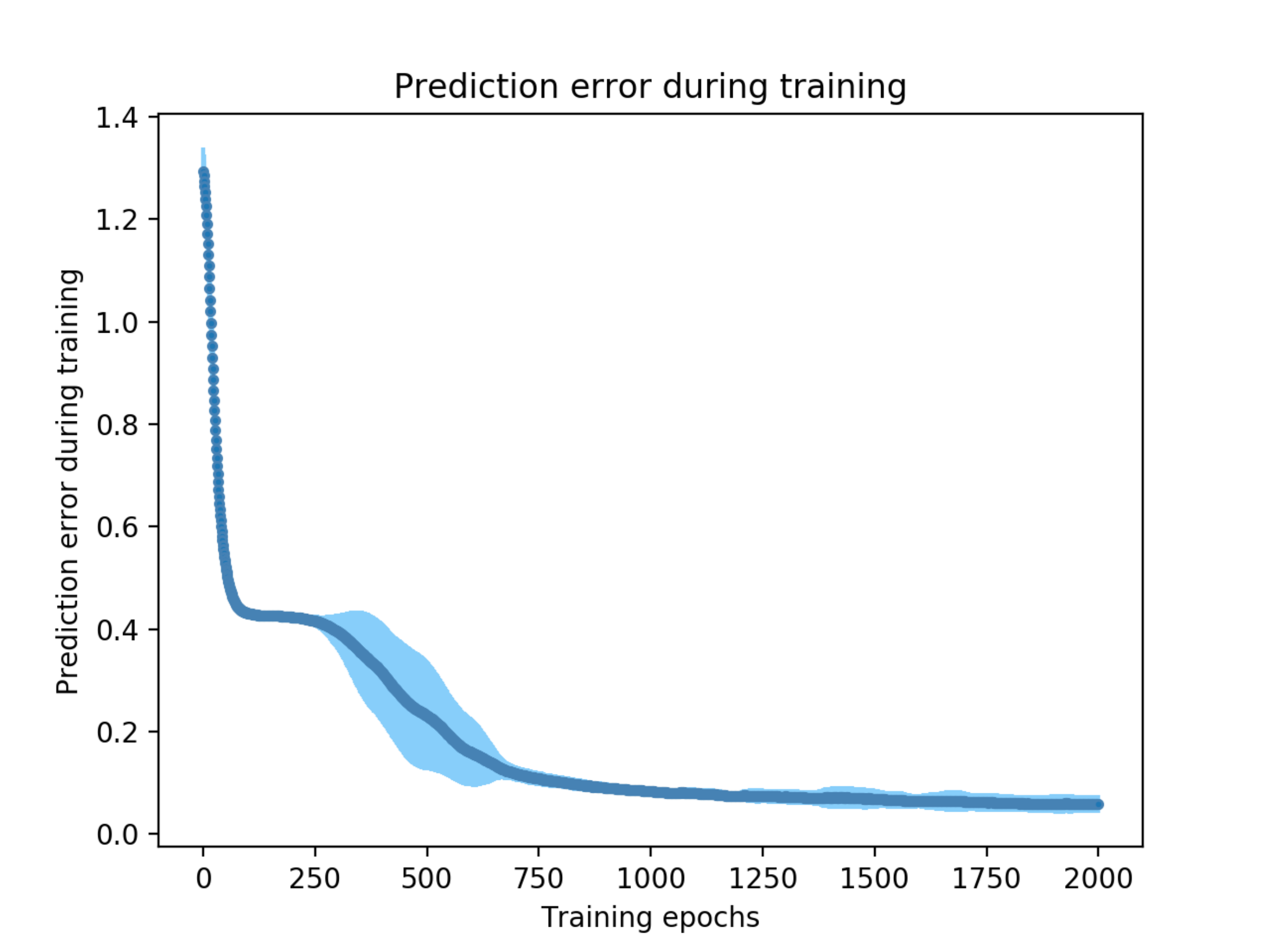}
	\caption{The average and standard deviation of the prediction error during training, averaged over ten differently initialized networks, when the gate is only open at the switches.}
	\label{fig:predErrSTD}
\end{figure}

Figure~\ref{fig:predErrSTD} shows the development of the average prediction error and its standard deviation with respect to the best performing network setup, that is, the one where the gate only opens as the switches.
As can be seen, the error first plateaus at a level of $~0.4$, which approximately corresponds to an identity mapping. 
It then rather reliably appears to find the gradient towards distinguishing the four functions over the subsequent training epochs, thus converging to an error level that corresponds to the lower-error boundary of the single-layer LSTM network with perfect context information (cf. Table~\ref{Tab:SingleLSTM_MLP}).

All the above-mentioned results were obtained using a fixed weight update frequency of 20 time steps, backpropagating the error 20 time steps into the past. 
Table~\ref{Tab:weightUpdate} shows the effect when the update frequency is changed.
In this case, the gradually changing surprise signal provides better results because in some of the runs, the network with the gate open at the context switch only fails to find the gradient down to the $0.05$ error niveau. 
The gradual opening and closing effectively increases the error flow into LSTMc, increasing the likelihood of convergence.
Thus, in the future a gradual change from weak, smooth surprise signals to strong and progressively more punctual surprise signals should be investigated further. 
Indeed such a surprise signal can be expected when derived from $LSTMf$ during learning \cite{Butz:2019}.

\begin{table}[b!]
  \centering
  \begin{scriptsize}
  \caption{Average training prediction error when a gradual surprise signal is provided while using different weight update frequency settings.}
   \label{Tab:weightUpdate}
{
\setlength\tabcolsep{0.4pt}
\newcolumntype{Y}{>{\centering\arraybackslash}X}
  \begin{tabularx}{\linewidth}{Y c c c c c c}
    \toprule
   \multirow{2}{*}{Weight update frequency} & \multicolumn {2}{p{2.85cm}}{\centering Fixed at 35} & \multicolumn {2}{p{2.85cm}}{\centering Random 20-50} & \multicolumn {2}{p{2.85cm}} {\centering Random 10-30} \\ \cmidrule(l{0pt}r{1pt}){2-3} \cmidrule(l{1pt}r{1pt}){4-5} \cmidrule(l{1pt}r{0pt}){6-7}
   & avg. error & stdev error & avg. error  & stdev error & avg. error  & stdev error \\
    \midrule
    Always closed & 0.365 & 0.070 & 0.428  & 0.083 & 0.345 & 0.078 \\
    Always open & 0.270 & 0.071 & 0.224 & 0.022 & 0.206 & 0.018  \\
    Open at context switch & 0.200 & 0.142 & 0.318 & 0.122 & 0.166 & 0.149 \\
    Gradually opened & \textbf{0.106} & 0.077 & \textbf{0.103} & 0.041 & \textbf{0.070} & 0.013 \\
    \bottomrule
  \end{tabularx}
}
  \end{scriptsize}
\end{table}

Please remember that in the experiments above the context switch provided to LSTMc switches earlier than the actual function event switch.
As a result, LSTMc can prepare for the switch but should only pass the information down when the event switch is actually happening.
This is accomplished by the GRU-like module.
Instead, when the surprise signal is provided to LSTMc and the GRU-like gate is always open, the error less reliably drops to the $0.05$ niveau, as shown in Table~\ref{Tab:avgErrStd}. 
On the other hand, when the contextual information was provided exactly in tune with the currently ongoing event to LSTMc -- opening the gate only at the switches still yielded a slightly better performance than when the gate was always open (cf. Table~\ref{Tab:avgErrStd}).

It is also worth mentioning that when we ran the architecture with an MLP (an MLPf module) as the function processing layer (instead of LSTMf), the error stayed on an average of $.42$, without any difference between the gating mechanisms (cf. Table~\ref{Tab:avgErrStd}). 
It thus appears that the gradient information from LSTMf is more suitable to foster the development of distinct contextual codes. 

\begin{table}[htb!]
\caption{Average prediction error when (i) the surprise signal is fed to LSTMc, whereby the GRU-like gate is always open (Surp. to LSTMc), (ii) the context information is provided to LSTMc exactly in tune with the function event (In-tune CI to LSTMc), and when an MLPf is used instead of an LSTMf (MLPf).}
\label{Tab:avgErrStd}
\begin{scriptsize}
{
\setlength\tabcolsep{0.4pt}
\newcolumntype{Y}{>{\centering\arraybackslash}X}
	\begin{tabularx}{\linewidth}{Y c c c c c c}
    \toprule
    {Input to LSTMc / } & 
    	\multicolumn{2}{p{2.85cm}}{\centering Surp. to LSTMc}  & 
    	\multicolumn{2}{p{2.85cm}}{\centering In-tune CI to LSTMc} & \multicolumn{2}{p{2.85cm}}{\centering MLPf}  \\
    	    	\cmidrule(l{0pt}r{1pt}){2-3} \cmidrule(l{1pt}r{1pt}){4-5} \cmidrule(l{1pt}r{0pt}){6-7}
    { Gate status} & avg. error & stdev  & avg. error  & stdev  & avg. error  & stdev  \\
    \midrule
    0 / Always closed & 0.2515  & 0.0678 & 0.310 & 0.080 & 0.4213 & 0.00164 \\
    1 / Always open  &  0.2280 & 0.0198 &  0.066  & 0.040 &  0.4215  & 0.00123 \\
    1 at c.s. / open at c.s. & 0.1031 & 0.0555 & 0.055 & 0.019 & 0.4211 & 0.00165 \\
    \bottomrule
  \end{tabularx}
 }
\end{scriptsize}
\end{table}

\begin{figure}[t!]
\begin{minipage}{\linewidth}
	       \centering
        \begin{subfigure}[b]{0.45\textwidth}
            \centering
            \includegraphics[width=\textwidth]{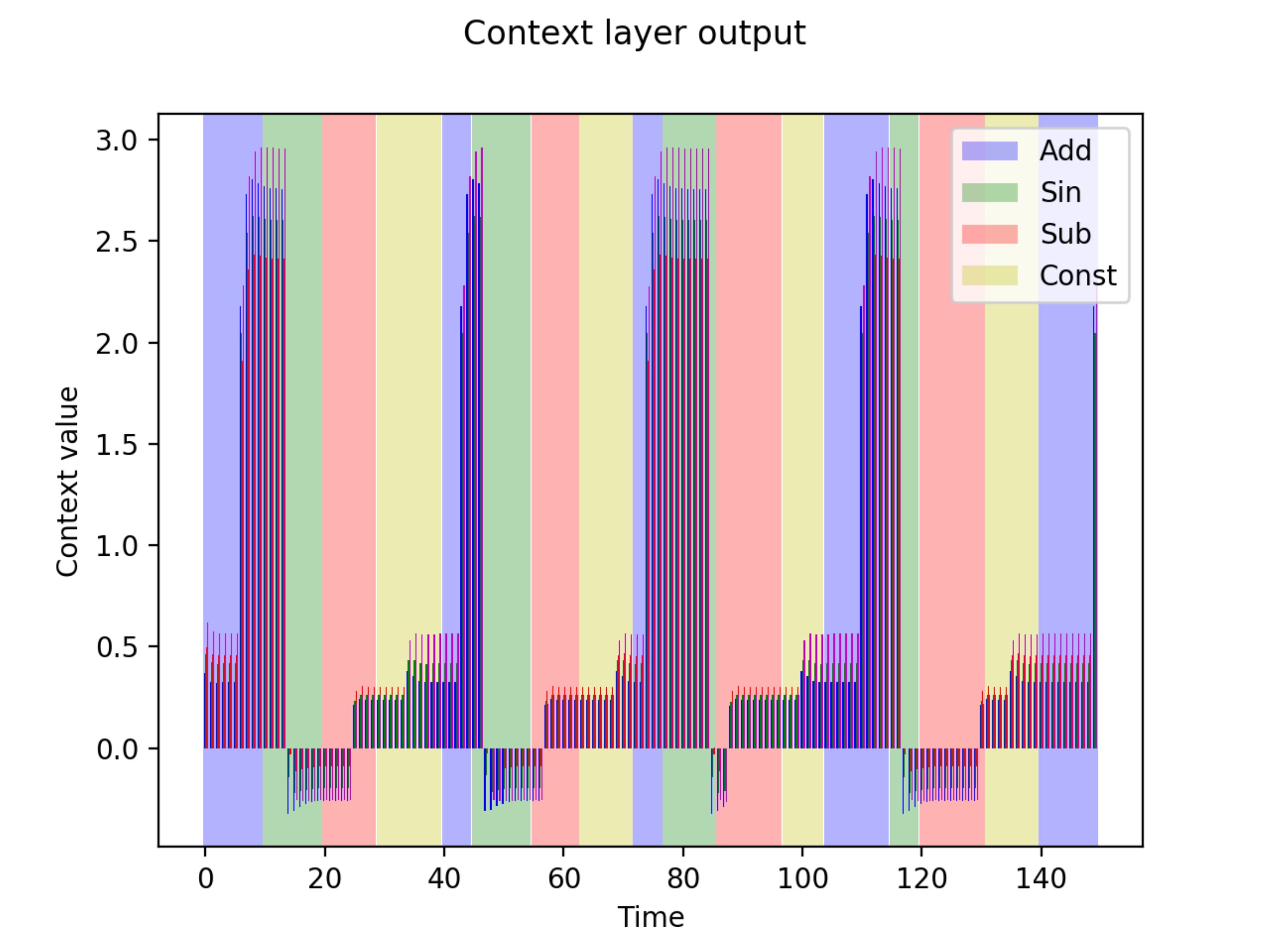}
            \caption[Gate open at switch]%
            {{\small Gate open at switch}}    
        \end{subfigure}
        \hfill
        \begin{subfigure}[b]{0.45\textwidth}   
            \centering 
            \includegraphics[width=\textwidth]{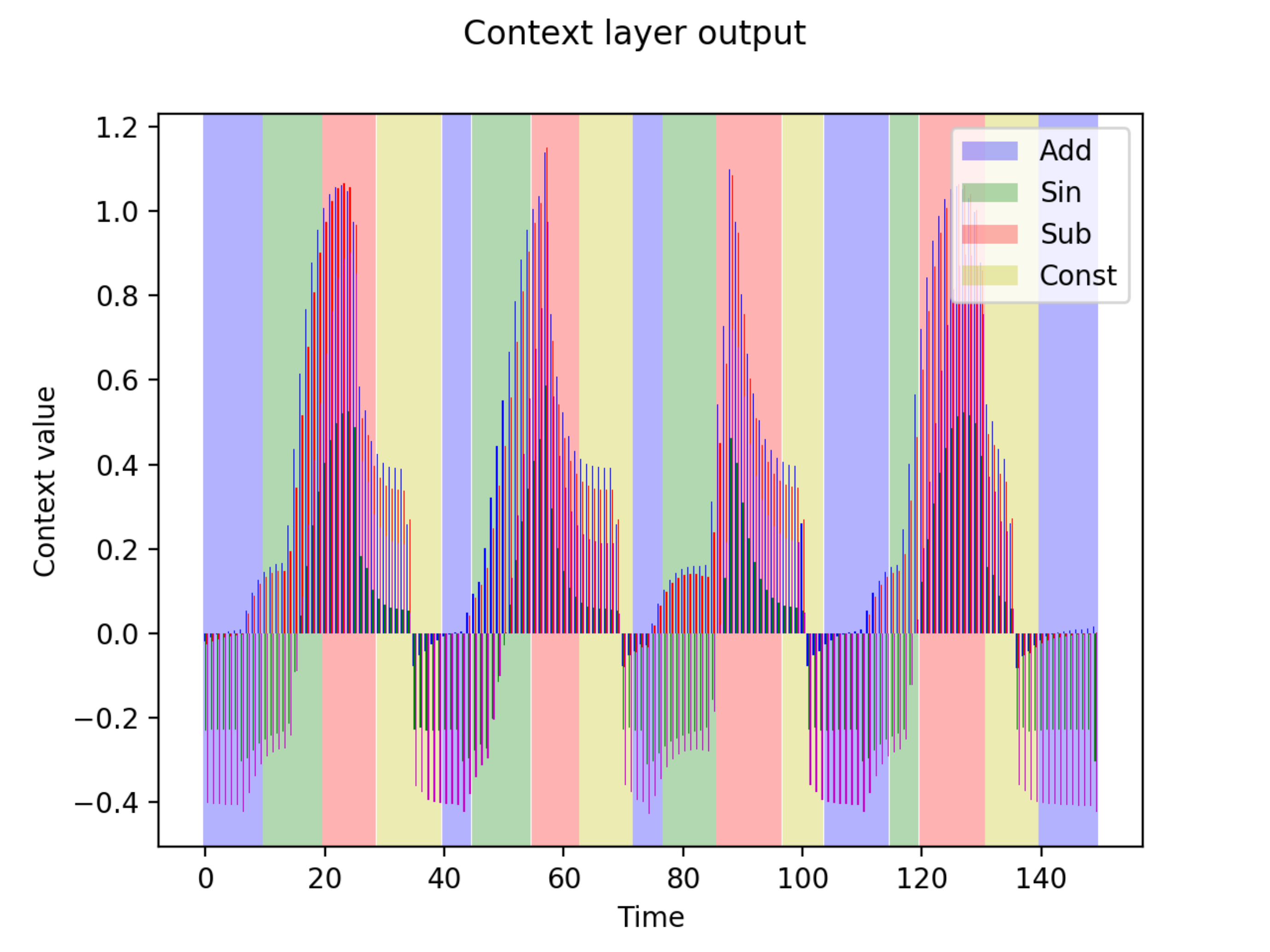}
            \caption[]%
            {{\small Gate always open}}    
        \end{subfigure}
        \quad
        \caption[ The context compressions produced by the context layer in the structure. The different background colors indicate the different contexts. ]
        {\small The context compressions produced by the context layer in the structure. The different background colors indicate the different contexts.} 
        \label{fig:contComp}
\end{minipage}

\begin{minipage}{\linewidth}
        \centering
        \begin{subfigure}[b]{0.45\textwidth}
            \centering
            \includegraphics[width=\textwidth]{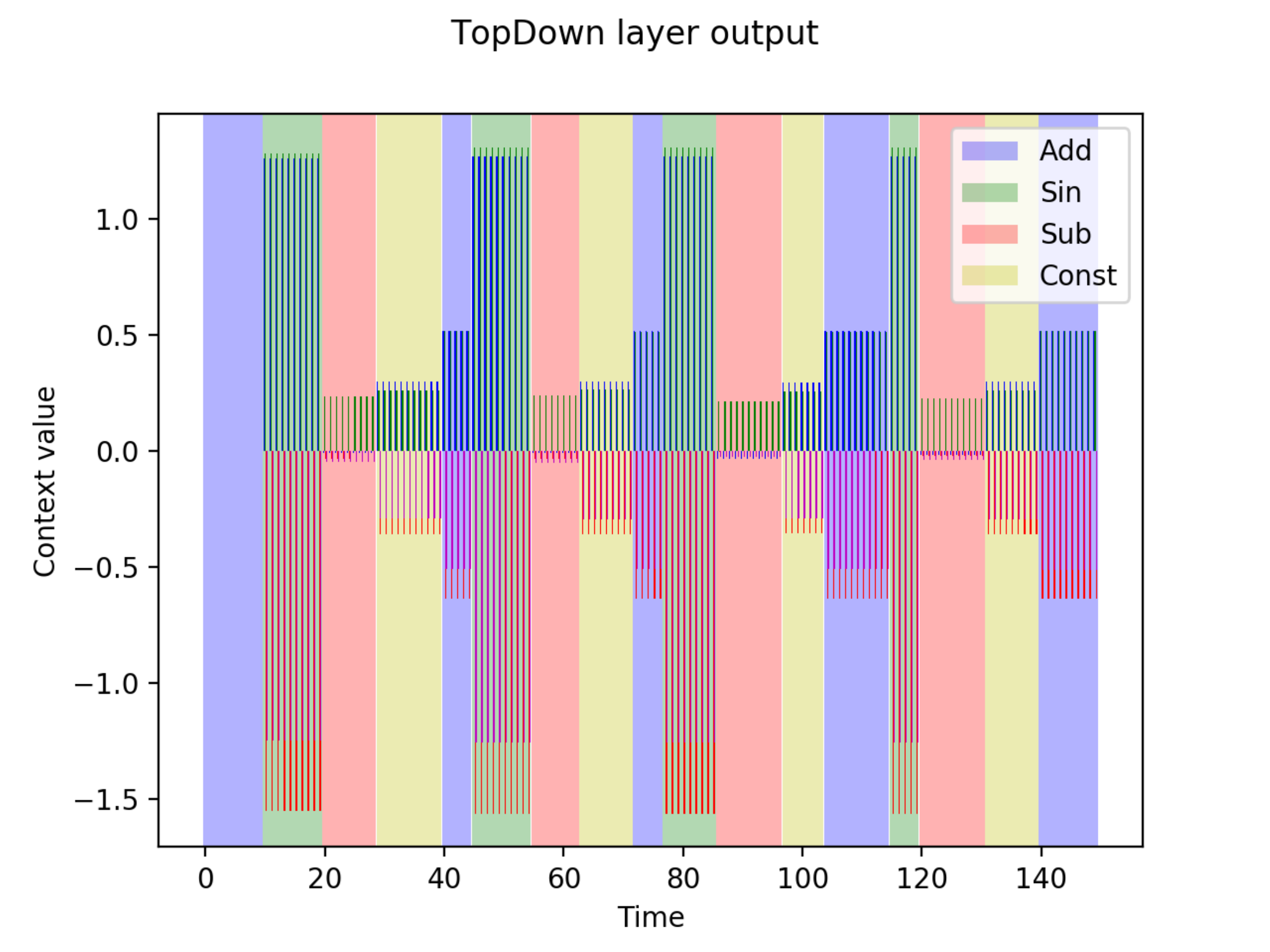}          \caption[Gate open at switch]%
            {{\small Gate open at switch}}    
        \end{subfigure}
        \hfill
        \begin{subfigure}[b]{0.45\textwidth}   
            \centering 
            \includegraphics[width=\textwidth]{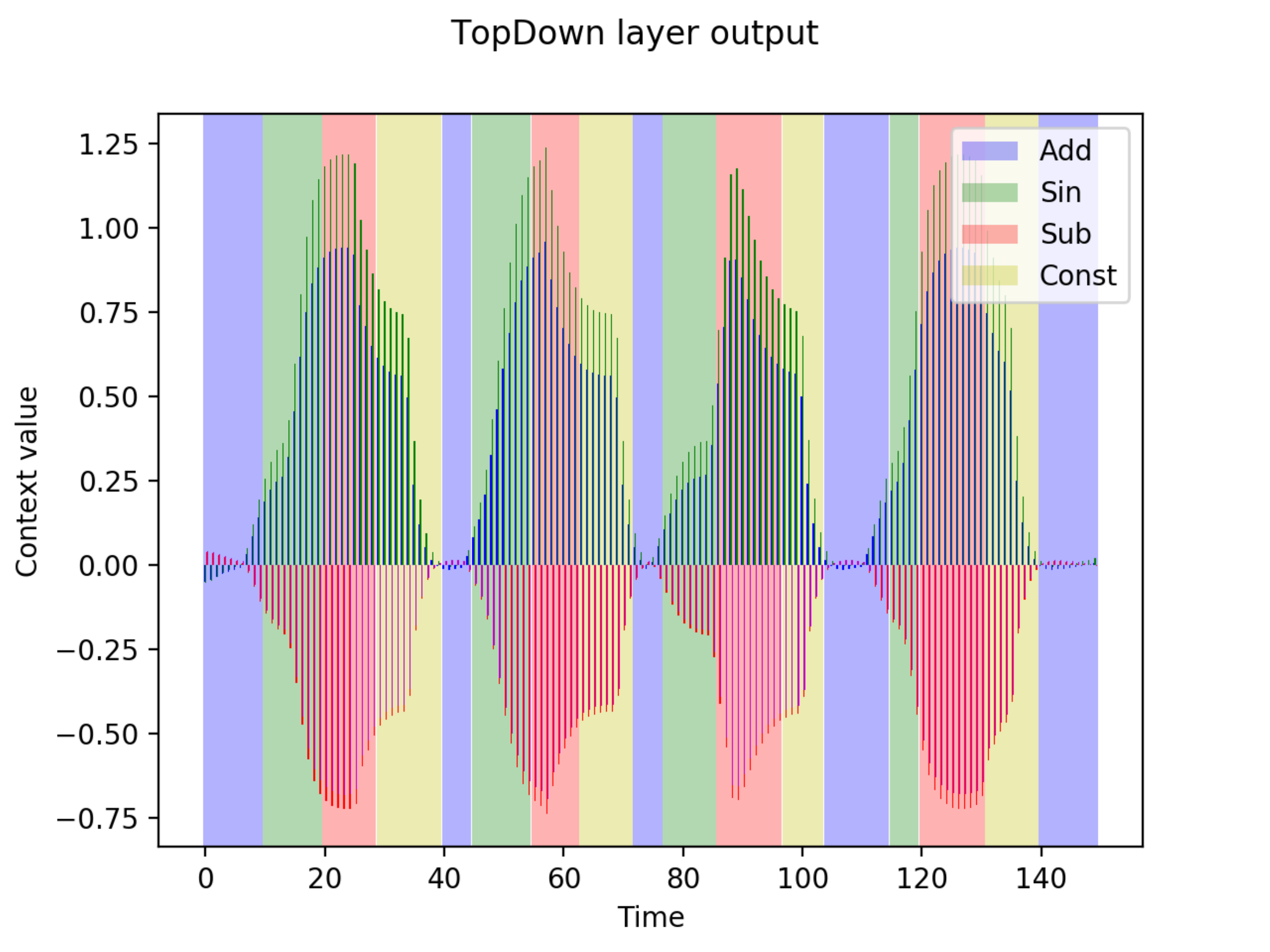}
            \caption[]%
            {{\small Gate always open}}    
        \end{subfigure}
        \quad
         \caption[ The context compressions provided by the GRU-like gating layer to the function processing layer. The different background colors indicate the different contexts. ]
        {\small The context compressions provided by the GRU-like gating layer to the function processing layer. The different background colors indicate the different contexts.} 
        \label{fig:topDownComp}
    \end{minipage}
\end{figure}

Finally, we took a closer look at the event-encoding compressions generated by the contextual layer and passed on by the GRU-like layer. 
Figure~\ref{fig:contComp} shows the context compression vector values produced by the deep context layer over time.
Figure~\ref{fig:topDownComp} shows the outputs of the GRU-like gating layer.
We can see stable compressions when the gate is only open at the switches.
When the gate is always open, the context also switches systematically but much more gradually and thus less suitably.

\begin{figure}[htb!]
	\centering
	\includegraphics[width=.9 \textwidth]{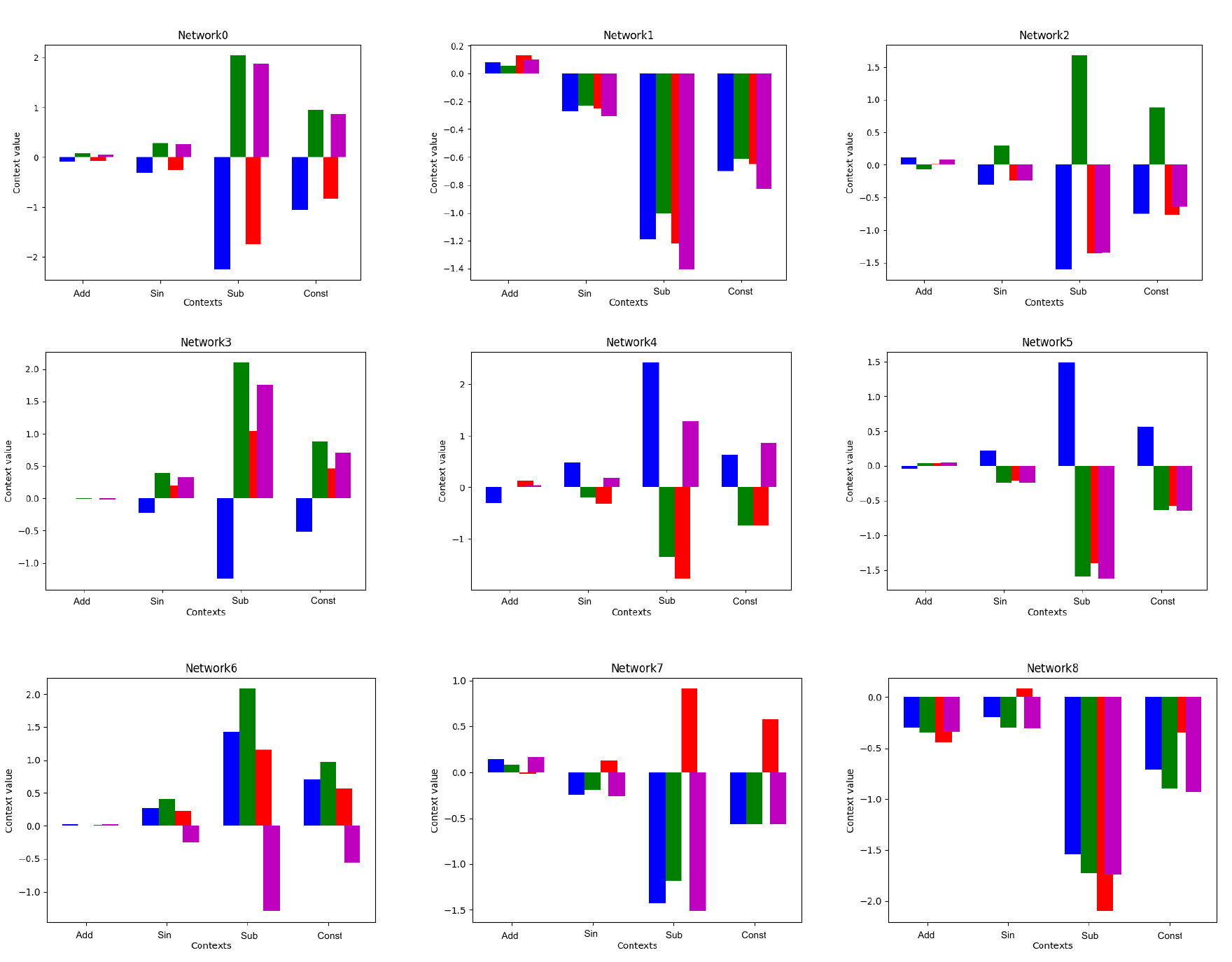}
	\caption{The different context compressions generated by the structure in nine differently initialized networks. Note that these compressions are different between different contexts in the same run.}
	\label{fig:avgTopDown}
\end{figure}

The results confirm that our surprise-processing modular architecture can clearly distinguish between the different contexts and the generated compressions vary between different networks. Figure~\ref{fig:avgTopDown} shows the context compressions for nine differently initialized networks.
It is interesting to note that the context-respective code for increasing is always close to zero, which is because the data always started with the increasing function at the beginning of an epoch and a network reset to zero activities. 
Moreover, it can be noted that, albeit clearly different, the constant function code lies somewhat in between the code for increasing and the code for decreasing. 
Finally, the sine function compression is distinct but also somewhat in between the increasing and constant function code (except for the in lower right graph). 
Further investigations with larger networks are pending to evaluate whether the sine function may be predictable more exactly with larger networks and whether the compression code from the GRU-like layer for the sine function may become more distinct from the others in that case.

\section{Discussion}
Motivated by recent theories on event-predictive cognition \cite{Butz:2016,zacks_event_2007,zacks_event_2020}, this paper has investigated how dedicated neural modules can be biased towards reliably developing even-predictive compressions. 

We have introduced a surprise-processing modular neural network architecture. 
The architecture contains a deep contextual layer, which generates suitable event-encoding compressions.
These compressions are selectively passed through a GRU-like top-down layer, depending on current estimates of surprise. 
If the surprise is low, then the same old compression is used. 
On the other hand, the larger the current surprise, the more of the current context compression is passed on to the function processing layer, effectively invoking an event transition.
As a result, the function processing layer predicts subsequent function values dependent on the currently active, compressed, top-down event-predictive signal.

Our surprise-processing modular architecture was able to generate best predictive performance when the GRU-like gating structures was opened only at or surrounding the event switch, mimicking the processing of a surprise signal. 
When the upcoming context information is provided in advance, the deep context layer does not only consider the currently ongoing event, but it also prepares the processing of the next one.
Thus, it is important that the gating top-down layer only passes the context compression when a new event actually starts.

Elsewhere, event-triggered learning was proposed for control, such that the system requests new information and the model is updated only when learning is actually needed \cite{solowjow_event-triggered_2018}.
To this end, our suggested structure shows that even when the context layer receives always the information regarding the actual ongoing event, the gate may still open only at the context switch, since this is the time point when new information needs to be passed to the actual event dynamics processing layer. 
As a result, the same prediction accuracy is achieved with a significantly more resource-efficient manner.

In future work, we will integrate surprise estimates from the LSTMf module directly, as previous analyzed elsewhere \cite{Butz:2019}.
Moreover, we intend to enhance the architecture further to enable it to predict event boundaries, whose detection initially correlates with measures of surprise \cite{hard_attention_2019}
Finally, the architecture will be combined with the REPRISE mechanism and scaled to larger problem domains, including robotic control and object manipulation tasks. 

%
%
%
\bibliographystyle{splncs04}
\bibliography{mybibliography}
%

\end{document}